# VIDEO OBJECT TRACKING AND ANALYSIS FOR COMPUTER ASSISTED SURGERY

Nobert Thomas Pallath[1] and Tessamma Thomas[2]

[1]Department of Electronics, Cochin University of Science & Technology, Kerala, India
nobert@cusat.ac.in
[2]Department of Electronics, Cochin University of Science & Technology, Kerala, India
tess@cusat.ac.in

*ABSTRACT*

*Pedicle screw insertion technique has made revolution in the surgical treatment of spinal fractures and spinal disorders. Although X- ray fluoroscopy based navigation is popular, there is risk of prolonged exposure to X- ray radiation. Systems that have lower radiation risk are generally quite expensive. The position and orientation of the drill is clinically very important in pedicle screw fixation. In this paper, the position and orientation of the marker on the drill is determined using pattern recognition based methods, using geometric features, obtained from the input video sequence taken from CCD camera. A search is then performed on the video frames after preprocessing, to obtain the exact position and orientation of the drill. An animated graphics, showing the instantaneous position and orientation of the drill is then overlaid on the processed video for real time drill control and navigation.*

## 1. INTRODUCTION

The technique of using pedicle screw into clinical practice by Roy-Camille et al. [1], [2] for spinal fixation was a major breakthrough in the field of spine surgery. The segments of the vertebral column are immobilized using surgical implants and bone grafts or otherwise called as internal fixation of the spine. This is useful in the treatment of fractured vertebrae, for the correction of spinal deformities in curvature like kyphosis, lordosis and scoliosis, treatment of back pain, for surgical management of neoplasms, degenerative diseases and stability [3].The procedure is to insert two screws into each vertebra to be fused. The angle of insertion of the pedicle screws is chosen in a manner, so as to avoid perforation of the pedicle which may cause damage to the spinal cord or roots [4]. This technique being successful in treating a wide variety of indication of spinal disorders has been widely adopted by orthopedic surgeons. Although the technique has many advantages, the placement of pedicle screw is a difficult procedure and has a high risk of misplacement. The impingement of nerve root alone has been found in 6.6% of all placements [4], [5]. Therefore accurate determination of the initial point of entry and the trajectory of screw insertion is extremely important.

Computer navigation systems serve as a useful aid in spine surgery [2], [6], [7]. Although pre-operative CT- imaging or registration is not required in fluoroscopy based navigation systems, CT based navigation systems have definite advantage with respect to precise preoperative planning using 3D visualization of patient anatomy [6]. Moreover, x-ray fluoroscopic technique has definite side effects, due to considerable radiation exposure to the patient and the surgical staff [1]. Also, it cannot be used during the entire screw insertion procedure due to possible spatial conflicts between C-frame, the surgeon and the surgical instruments [5]. Surgical robots are voluminous and occupy too much of the operating room space [5], [8]. Registration and immobilization are two key issues in robot assisted surgery [9]. Also, commercial surgical robots are extremely costly. Although these methods claim over 90 percent accuracy, their use is limited to few large research hospitals [4], [5]. In this paper, we present a computer assisted method with low instrumentation cost and high precision using real time video processing and computer graphics.

## 2. MATERIALS AND METHODS

The method developed is based on real time processing of the video grabbed using the experimental setup, consisting of cadaveric dry human vertebra, phantom model of the vertebra, micro motor drill, Cohu DSP 3600 cameras, workstation computer with matrox morphis frame grabber, in a created surgical environment. The optimum distance, position, yaw, pitch and roll of the camera are fixed. The cameras are placed at a position considering the entire surgical setup like patient position, lighting and also without causing any obstruction to the surgeon and the entire surgical set up. The workstation computer is arranged with the monitor at a viewable distance.

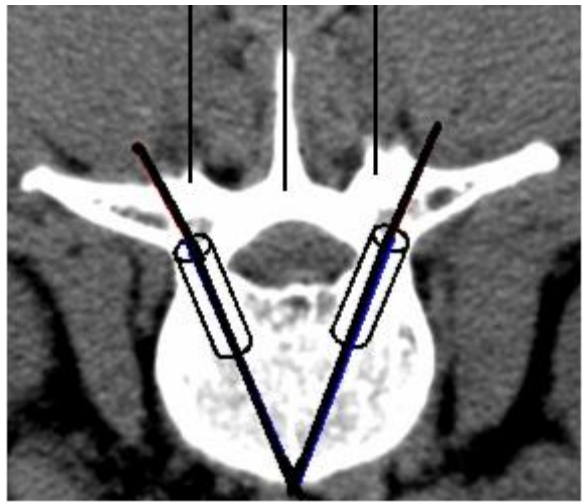

Figure1. An example of the reference image

### 2.1 Pre-operative Planning

Pre-operative planning is an important step in the procedure which involves the analysis and measurement of the pedicle parameters viz., width, height and orientation. The pre-operative axial CT- image of the spine is used for this purpose. The step involves identification of the vertebrae where the pedicle screw is to be inserted, selection of the appropriate representative image of the vertebra, marking the landmark points on the selected image and computing the parameters of the pedicle like width and height. The process is done for both the vertebrae used for fusing. 3D doctor software is used for vertebra modeling and measurements [5]. A one inch square marker with a unique geometric shape is designed by considering the shapes of all background objects so as to avoid ambiguity and false detection during object search. The marker is fixed centrally on the body of the micro motor drill. An alternate method of fixing the marker on the drill owl is also used for tracking the pedicle screw. The axis of the drill or drill owl passes through the centroid of the marker.

The axial CT image of the candidate vertebra consists of eight or nine slices at a separation of 2 to 3 mm. The fourth or fifth slice is the best representative image [5]. This image provides a clear picture about the pedicle dimensions, from normal anatomy. The image as shown in the figure1 is used to determine the pedicle width, angle and relationship with other anatomical structures. A vertical line is drawn through the middle of the transverse process and equidistant lines from the central lines drawn in each of the spineous process as shown in figure1, aid in the registration step [5]. Registration of the CT image and the actual vertebra is done by overlaying. Two lines drawn through the centre of the pedicle area from the lamina to the vertebra body as shown in figure1, displays the ideal reference path for pedicle screw insertion [5].

The anatomy of the pedicle shows that, it has a non–uniform cylindrical shape, with varying diameter across its length [5]. Graphical cylinder plotted with diameter, fixed using minimum width of the pedicle area as shown in figure1, aids in visualization of the trajectory and tracking of the pedicle screw during insertion [5].

## 2.2 Camera Calibration

Relationship between pixel coordinates and real world coordinates is established using camera calibration. A dot pattern grid is used to map pixel coordinates to real world coordinates, for accurate analysis and measurement of the drill position and orientation. A square grid pattern is used, for detecting perspective distortions due to camera lens. The mapping physically corrects image distortions, viz. non unity aspect ratio distortion, rotation distortion, perspective distortion, pincushion distortion and barrel type distortion. The results are returned in real world units, which automatically compensates for any distortions in the image. A calibration object is used to hold the defined mapping and used to transform pixel coordinates or results to their real world equivalents.

Using the theorem of intersecting lines [10], the computational model of the pinhole camera model is denoted by:

$$\begin{pmatrix} u \\ v \end{pmatrix} = \frac{f}{z} \begin{pmatrix} x \\ y \end{pmatrix} \qquad [10]$$

where, x, y, z the coordinates of a scene point in the 3D coordinate system whose origin is the projection center and u, v denote the image coordinates. The parameter f is known as the camera constant; it denotes the distance from the projection center to the image plane.

From figure 2,

$$\begin{pmatrix} u \\ v \end{pmatrix} = \frac{f}{D} \begin{pmatrix} x \\ y \end{pmatrix} \qquad (1)$$

Where $z \approx D$ the axial distance.

Also

$$D = \sqrt{V^2 + H^2}$$

Therefore,

$$u = \frac{f}{\sqrt{V^2 + H^2}} \times x \qquad (2)$$

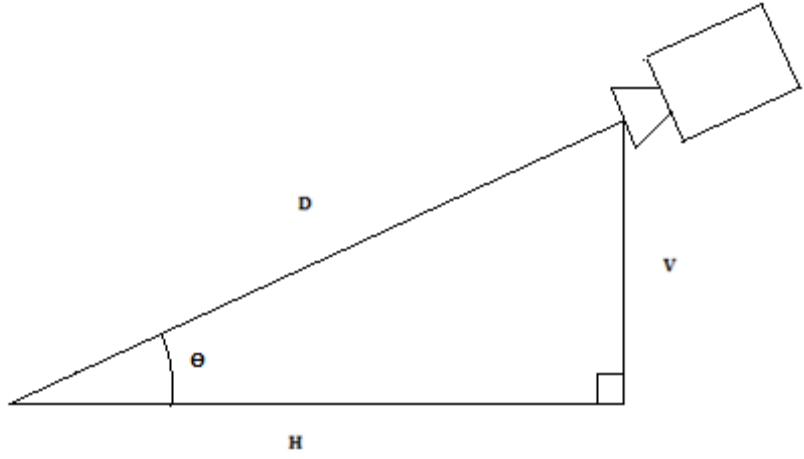

Figure 2. Camera placement and distances

From equation (2),

$$(u_1 - u_2) = \frac{f}{\sqrt{V^2+H^2}} \times (x_1 - x_2)$$

Therefore,

$$\frac{(u_1-u_2)}{(x_1-x_2)} = \frac{f}{\sqrt{V^2+H^2}}$$

Let $P_d$ be the Pixel distance with respect to the object displacement $R_d$. Therefore, the Euclidean distance between any two pixel positions is,

$$P_d = P_{d1} - P_{d2} = \sqrt{(u_1-u_2)^2 + (v_1-v_2)^2}$$

The corresponding object displacement,

$$R_d = R_{d1} - R_{d2} = \sqrt{(x_1-x_2)^2 + (y_1-y_2)^2}$$

The ratio,

$$\frac{P_d}{R_d} = \frac{\sqrt{(u_1-u_2)^2+(v_1-v_2)^2}}{\sqrt{(x_1-x_2)^2+(y_1-y_2)^2}} \qquad (3)$$

From equation (2),

$$\frac{\sqrt{(u_1-u_2)^2+(v_1-v_2)^2}}{\sqrt{(x_1-x_2)^2+(y_1-y_2)^2}} = \frac{f}{\sqrt{V^2+H^2}}$$

Therefore, $\dfrac{P_d}{R_d} = \dfrac{f}{\sqrt{V^2+H^2}}$ \qquad (4)

Or,

$$f = \frac{P_d}{R_d} \times \sqrt{V^2 + H^2} \qquad (5)$$

Using a set of object points { $(x_1,y_1), (x_2,y_2),\ldots (x_n,y_n)$ }, the corresponding image points { $(u_1,v_1), (u_2,v_2), \ldots(u_n,v_n)$ }, are obtained using the camera and the ratio 'Pd/Rd' is found out from (3). Next, the value of V and H are measured after fixing the camera. Knowing the ratio '$P_d/R_d$', V and H, the value of $f$ is found out using (5). Now, knowing $f$, V and H, the value of '$R_d$' can be found out for every measured '$P_d$'.

## 2.3 Registration and Surgery

After surgical exposure of the spine, one needle is placed in the middle of the superior articular process and two needles are placed, on the spineous process, at distances measured during the pre-operative planning phase [5]. By overlaying the transparent reference image, with lines drawn as mentioned in section [2.1], over the video and adjusting the focus and zoom of the camera, the three needles in the video, are exactly made to coincide with the three vertical lines, plotted on the reference image. At this stage, the dimensions of the objects in both the images match, which finalizes the registration process. Now, the drill is positioned with its burr exactly placed at the entry point. Using computer graphics, the cylinder and its axis, with the required height and diameter, measured during the pre-planning phase, are created.

Square marker with 2.5 cm $\times$ 2.5 cm dimension, having a unique geometric shape is designed by considering the shapes of all background objects, so as to avoid ambiguity and false detection. Marker is fixed centrally on the body of the micro motor drill, so as to face the camera. An alternate method is to fix the marker, on the drill owl, so as to track the pedicle screw. The axis of the drill or drill owl passes through the centroid of the marker. The video of the drill, with the marker fixed centrally on its body is grabbed and processed in sequential frames.

The procedure begins by correcting the orientation of the drill so that, it correctly enters the pedicle canal and the vertebral body. The orientation of the drill is same as the marker orientation. Now, the path of the drill is tracked during insertion, to ensure that it does not go beyond the walls of the pedicle canal or pierce the vertebra body. The method is to search the marker, using edge extraction to get the geometric features of the marker. The search is performed and results are displayed, based on calibration. The algorithm uses edge based geometric features of the models and the target, to establish match. Gradient method is used, for extracting object contours. An object contour is a type of edge that defines the outline of the objects in an image. The edges extracted from the video frame are used to form the image's edge map, which represents how the image is defined as a set of edges. The feature calculations are performed using the image's edge map. The edge finding method uses operations that are based on differential analysis, where edges are extracted by analyzing intensity transitions in images. Edges are extracted in three basic steps. First, a filtering process provides an enhanced image of the edges, based on the computations of the image's derivatives. Second, detection and thresholding operations determine all pertinent edge elements, or edgels from the image. Third, neighboring edgels are connected to build the edge chains and features are calculated for each edge. The enhanced image of the object contours is obtained by calculating gradient magnitude of each pixel in the image.

First order derivatives of a digital image are based on various approximations of the 2D gradient. The gradient of an image f(x, y) at the location (x, y) is defined as the vector [11]:

$$\nabla f = \begin{pmatrix} Gx \\ Gy \end{pmatrix} = \begin{pmatrix} \frac{\partial f}{\partial x} \\ \frac{\partial f}{\partial y} \end{pmatrix} \qquad (6)$$

The gradient magnitude is calculated at each pixel position, from the image's first derivatives. It is defined as [11]:

$$\text{Gradient Magnitude} = \text{mag}(\nabla f) = \sqrt{Gx^2 + Gy^2} \qquad (7)$$

An edgel or edge element is located at the maximum value of the gradient magnitude over adjacent pixels, in the direction defined by the gradient vector. The gradient direction is the direction of the steepest ascent at an edgel in the image, while the gradient magnitude is the steepness of that ascent. Also, the gradient direction is the perpendicular to the object contour.

The marker with the unique geometric shape is fixed as the search model. The search of instances of models in the sequence of video frames is performed. The match between the model and its occurrences in the target image is determined using the values of 'score' and 'target score'. The score is a measure of active edges of the model found in the occurrence, weighted by the deviation in position of these common edges. The model scores are calculated as follows.

Score = Model coverage × (1- (Fit error weighing factor × Normalized Fit Error))

Target score = Target coverage × (1- (Fit error weighing factor × Normalized Fit Error))

The model coverage is the percentage of the total length of the model's active edges, found in the target image. 100% indicates that, for each of the model's active edges, a corresponding edge was found in the occurrence. The target coverage is the percentage of the total length of the model's active edges, found in the occurrence, divided by the length of edges present within the occurrence's bounding box. Thus, a target coverage score of 100 % means that, no extra edges were found. Lower scores indicate that, features or edges found in the target are not present in the model. The fit error is a measure of how well the edges in the occurrence, correspond to those of the model. The fit error is calculated as the average quadratic distance, in pixels or calibrated units, between the edgels in the occurrence and the corresponding active edges in the model.

$$\text{Fit error} = \frac{\sum_{\text{all common pixels}} [(\text{error in x})^2 + (\text{error in y})^2]}{\text{Number of common pixels}}$$

A perfect fit gives a fit error of 0.0. The fit error weighing factor (between 0.0 – 100.0) determines the importance to place on the fit error when calculating score and target score. An acceptance level is set for both the score and target score.

A graphical line, showing the position and orientation of the marker on the drill, is constructed within the graphical cylinder using line drawing technique in computer graphics, and is displayed in real time, by using the position and orientation of the centroid of the marker and drawing the results, in the display's overlay buffer non-destructively. The line is displayed within the graphical cylinder with its axis at exact inclination as that of the axis of the pedicle canal and its dimensions exactly same as that of the pedicle canal, constructed earlier using computer graphics. The graphical results display the position and orientation of the drill and are used for real time drill control and navigation. Positional results and audio-visual alerts are used to prevent boundary violation, which can lead to pedicle wall perforation. An interactive GUI and real time video display, with real time graphical overlay is built for ease of access, for viewing position and orientation of the drill or pedicle screw during insertion.

## 3. EXPERIMENTAL SETUP AND RESULTS

### 3.1 Camera Calibration

Table 1. Camera calibration for selected distances

| Camera position w.r.t. object in world co-ordinates | | | | Marker displacement in world co ordinates | Marker displacement in the image co-ordinates | | | | | |
|---|---|---|---|---|---|---|---|---|---|---|
| Calculated Axial distance D (cm) | Vertical Distance V (cm) | Horizontal Distance H (cm) | Calculated angle between horizontal axis and camera axis $\theta$ (deg.) | Real world distance corresponding to marker displacement Rd(cms) | X1 Co-ordinate (pixel units) | Y1 Co-ordinate (pixel units) | X2 Co-ordinate (pixel units) | Y2 Co-ordinate (pixel units) | Pixel distance corresponding to marker displacement Pd (pixel units) | Ratio Pd/Rd |
| 86.0 | 59.0 | 62.6 | 43.3 | 2.0 | 325.0 | 224.0 | 335.0 | 209.0 | 18.0 | 9.2 |
| 104.3 | 59.0 | 86.0 | 34.5 | 2.0 | 353.0 | 225.0 | 361.0 | 214.0 | 13.6 | 6.9 |
| 125.5 | 59.0 | 110.8 | 28.0 | 1.9 | 355.0 | 137.0 | 363.0 | 128.0 | 12.0 | 6.3 |
| 150.1 | 59.0 | 138.0 | 23.1 | 1.9 | 349.0 | 48.0 | 354.0 | 38.0 | 11.2 | 5.8 |

The camera calibration was done, using dot pattern grid for different camera distances, relative to the object position. Calibration was also done using marker displacement for different camera distances and the ratio 'Pd/Rd' was calculated using (3). The calculated values of 'Pd/Rd' for selected distances are shown in table 1. It shows that, as the distance between the object and camera increases, the value of the ratio, 'Pd/Rd' decreases.

### 3.2 Determination of Calibration Accuracy

Next, to evaluate the accuracy of the calibration done, measurements were taken, by fixing the camera position and measuring the distances 'V' and 'H' of the camera, relative to the object. The ratio 'Pd/Rd' is found by using (3). Equation (5) is used to find out the values of '$f$', using the values of 'V', 'H' and 'Pd/Rd', using (3) .The values of '$R_d$' for different marker positions are shown in table 2. It shows that, by fixing the values of '$f$', 'V' and 'H', marker displacement '$R_d$' can be found out from the corresponding measured value of pixel displacement '$P_d$' with a minimum precision of ± 0.5mm.

After fixing the camera position and orientation, the calibration is finalized for mapping pixel coordinates to real world coordinates, to correct image distortions and for precise analysis and measurement of the drill position and orientation.

### 3.3 Evaluation using Phantom Model of the Vertebra

The new technique was evaluated, by inserting the drill into the pre-determined point, of the transparent phantom model of the human vertebra, using computer assistance. Three needles were inserted into the landmark points on the phantom vertebra, as mentioned in section [2.3].

The focus and zoom of the camera were adjusted so that, the three needles in the video were exactly made to coincide with the three vertical lines plotted on the reference CT image to complete the registration process. The graphical cylinder was drawn, with its axis at an inclination, exactly same as that of the pedicle canal, obtained from the reference CT image of the vertebra. The orientation of the axis of the cylinder was estimated, with respect to the three vertical lines drawn in the reference CT image of the vertebra.

Table 2. Evaluation of Calibration Accuracy

| Marker displacement in world co-ordinates | | | | | Marker displacement in the image co-ordinates for camera position relative to object D = 95.2 cm, V = 56.0cm and H = 77.0 cm | | | | | | | |
|---|---|---|---|---|---|---|---|---|---|---|---|---|
| X1 (cm) | Y1 (cm) | X2 (cm) | Y2 (cm) | Real world distance corresponding to marker displacement $R_d$ (cms) | X1 Co-ordinate (pixel units) | Y1 Co-ordinate (pixel units) | X2 Co-ordinate (pixel units) | Y2 Co-ordinate (pixel units) | Pixel distance corresponding to marker displacement $P_d$ (pixel units) | Ratio $P_d/R_d$ | $f = \frac{P_d}{R_d} \times \sqrt{V^2 + H^2}$ | Calculated $R_d = P_d \times \frac{\sqrt{V^2 + H^2}}{f}$ using $f = 1028.2$ |
| 16.4 | 10.6 | 15.0 | 9.4 | 1.9 | 422.0 | 429.0 | 406.0 | 441.0 | 20.0 | 10.8 | 1028.2 | 1.85 |
| 15.0 | 11.9 | 14.0 | 10.2 | 1.9 | 405.0 | 416.0 | 393.0 | 433.0 | 20.8 | 10.8 | 1025.4 | 1.93 |
| 12.0 | 13.0 | 11.9 | 11.8 | 1.2 | 368.0 | 406.0 | 372.0 | 418.0 | 12.6 | 10.9 | 1034.2 | 1.16 |
| 7.7 | 11.8 | 9.0 | 10.4 | 1.9 | 311.0 | 424.0 | 327.0 | 437.0 | 20.6 | 10.8 | 1024.9 | 1.92 |
| 6.8 | 10.8 | 8.4 | 9.7 | 1.9 | 300.0 | 435.0 | 318.0 | 445.0 | 20.6 | 10.8 | 1024 | 1.91 |

A square marker with 2.5 cm × 2.5 cm dimension, having a unique geometric shape was designed. The marker was fixed centrally on the body of the micro motor drill, facing the camera. The axis of the drill passes through the centroid of the marker so that, orientation of the drill is same as the marker orientation. Figure 3, shows the setup.

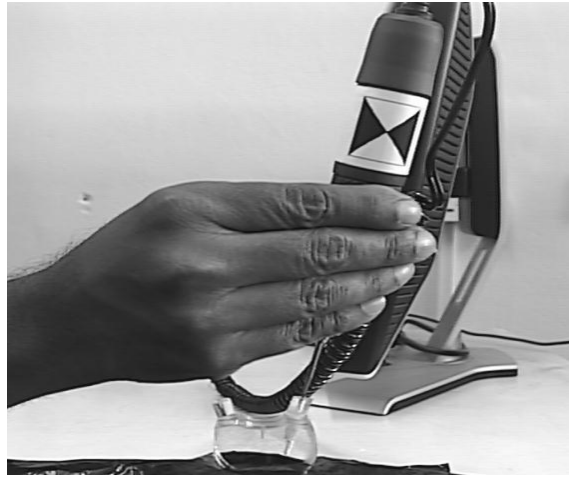

Figure 3. Marker fixed on the micro motor drill

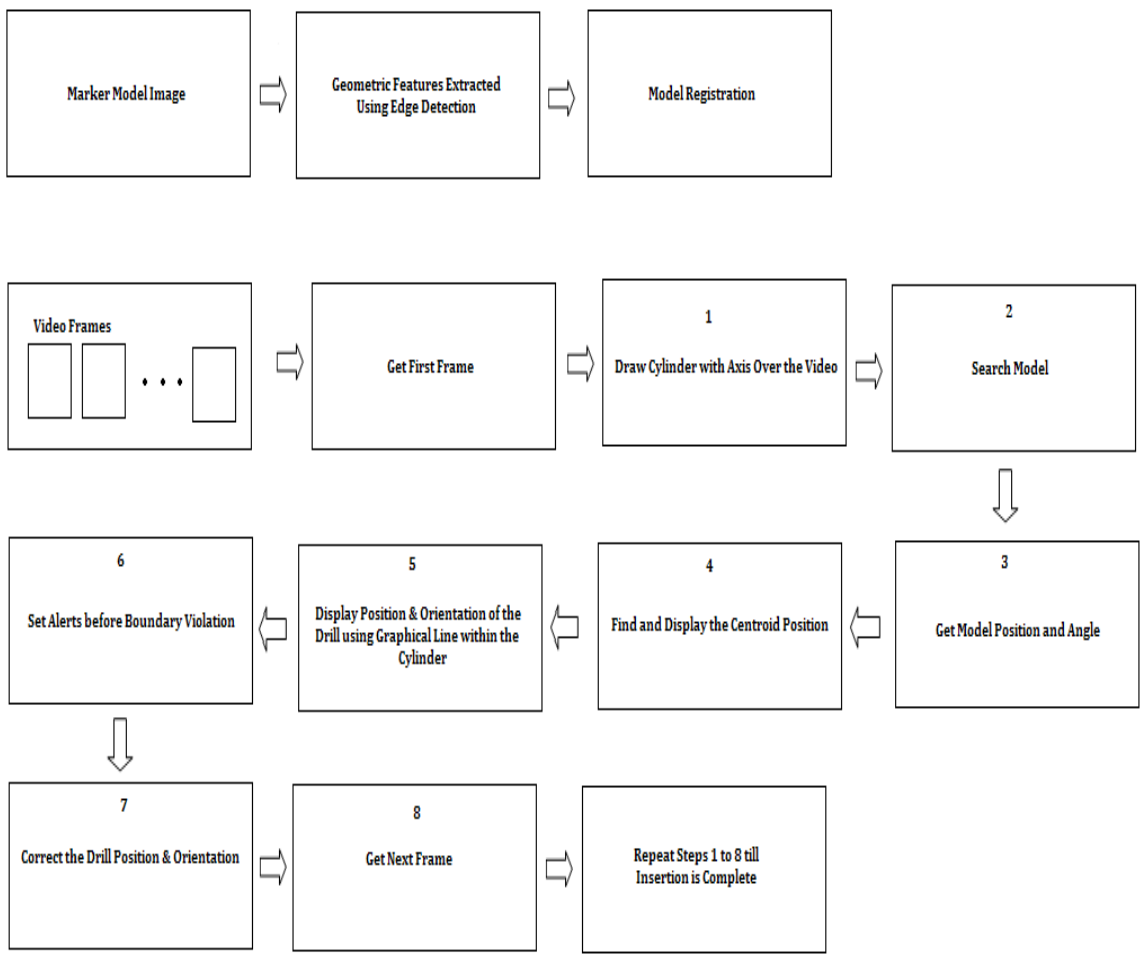

Figure 4. Video object tracking for CASS

The video of the drill, with the marker fixed centrally on its body was grabbed and processed in sequential frames. The search of instances of marker models in the sequence of video frames was performed. The centroid of the marker model was found out in each frame of the video. A graphical line, showing the position and orientation of the centroid of the marker on the drill, was constructed within the graphical cylinder using line drawing technique in computer graphics, and was displayed in real time, by using the position and orientation of the centroid of the marker and drawing the results, in the display's overlay buffer non-destructively. Then, the drill was positioned with its burr exactly placed at the entry point on the phantom vertebra. The orientation of the drill was corrected so as to correctly enter the pedicle canal. The orientation of the drill should be the same as the marker orientation. Then, the path of the drill was tracked during insertion, so that it neither goes beyond the walls of the pedicle canal nor pierces the vertebra body. The trajectory of the burr or tip of the drill was viewed in real time, by observing the movement of the graphical line within the cylinder. The depth of insertion was estimated by viewing the movement of the graphical line. Figure 4, illustrates the procedure of Computer Assisted Spine Surgery (CASS).

### 3.4 Results

A new algorithm was developed for CASS. The maximum search time required was only 11.5 mS, which is good for real time performance. A user friendly GUI with provision for testing the camera, loading marker image, tracking object, display positional results and search time per frame has been developed. Animated graphical overlay over real time video has been developed using computer graphics, for ease of access and viewing the position and orientation of the drill or pedicle screw during insertion. An error of only ±0.5 mm was observed between real and calculated drill positions. Audio and visual alerts, together with positional results, aid in precise drill control and navigation.

## CONCLUSIONS

A real time video based surgical navigation system for pedicle screw insertion was developed. The system provides precise drill orientation correction in real time. The trajectory of insertion of the drill or the pedicle screw is displayed in real time and provides an aid to the surgeon, to insert the screw precisely. The system developed is cost effective and has a good precision of ± 0.5 mm and search time of 11.5 mS. The instrumentation required is simple so that, handling the system is fairly easy.

## REFERENCES

[1]   Lutz P. Nolte, M.A Slomczykowski, Uirich Berlemann, Matthias J. Strauss, Robert Hofstetter, Dietrich Schlenzka, Timo Laine, Teija Lund, " A new approach to computer aided spine surgery: fluoroscopy based surgical navigation", *Eur Spine J (2000) 9 : S78-S88,* Springer–Verlag 2000

[2]   Robert W. Gaines, JR., M.D., " The Use of Pedicle-Screw Internal Fixation for the Operative Treatment of Spinal Disorders", *The Journal of Bone and Joint Surgery Vol. 82-A, No. 10, October 2000*

[3]   Muris Mujagi´c, Howard J. Ginsberg, and Richard S. C. Cobbold, "Development of a Method for Ultrasound-Guided Placement of Pedicle Screws", *IEEE Transactions on Ultrasonics, Ferroelectrics, and Frequency Control,* Vol. 55, no. 6, June 2008

[4]   Moshe Shoham*,* Member, IEEE, Michael Burman, Eli Zehavi, Leo Joskowicz*,* Senior Member, IEEE, Eduard Batkilin, "Bone-Mounted Miniature Robot for Surgical Procedures: Concept and Clinical Applications and Yigal Kunicher" *, IEEE Transactions On Robotics And Automation,* VOL. 19, NO. 5, October 2003

[5] Tessamma Thomas, Dinesh Kumar V.P., P.S John, Antony Joseph Thoppil, James Chacko, " A Video Based Tracking System for Pedicle Screw Fixation", *Proc. of 4th International Conference on Computer Science and its Application ( ICCSA -2006 ) san Diego, California, USA,* June 2006

[6] Yuichi Tamura, NobuhikoSugano, ToshihikoSasama,YoshinobuSato, Shinichi Tamura, Kazuo Yonenobu, Hideki Yoshikawa, Takahiro Ochi, "Surface-based registration accuracy of CT-based image-guided spine surgery", *Eur Spine J (2005) 14:* 291–297

[7] T. Laine,T. Lund,M. Ylikoski,J. Lohikoski,D. Schlenzka, "Accuracy of pedicle screw insertion with and without computer assistance : a randomised controlled clinical study in 100 consecutive patients", *Eur Spine J (2000) 9 :235–240, Springer-Verlag 2000*

[8] T. Ortmaier, H. Weiss, U. Hagn, M. Grebenstein, M. Nickl, A. Albu-Sch¨affer,C. Ott, S. J¨org, R. Konietschke, Luc Le-Tien, and G. Hirzinger, "A Hands-On-Robot for Accurate Placement of Pedicle Screws", *IEEE International Conference on Robotics and Automation May 2006*

[9] Jon T. Lea, Dane Watkins, Aaron Mills, Michael A. Peshkin,Thomas C. Kienzle III, S. David Stulberg "Registration and immobilization in robot-assisted surgery" conference proceedings of the *First International Symposium on Medical Robotics and Computer Assisted Surgery, Pittsburgh, PA*, Sept.1994

[10] Pedram Azad, Tilo Gockel, Rüdiger Dillmann, *Computer Vision Principles and Practice* 1st Edition, Elektor International Media BV 2008

[11] Rafael C. Gonzalez, Richard E. Woods, *Digital Image Processing*, second edition, PrenticeHall. 2004

**Authors**

**Nobert Thomas Pallath** received M.Sc. (Electronics) from Cochin University of Science and Technology, Cochin-22, India. He has been working as Assistant Professor, in the Department of Electronics, W.M.O College, Wayanad, Kerala, India. Currently, he is doing Ph.D. in the Department of Electronics, Cochin University of Science and Technology, India. His research interest includes image processing, computer vision and signal processing.

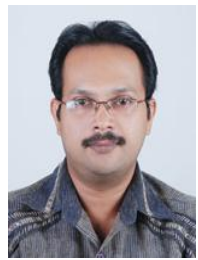

**Dr. Tessamma Thomas** received M. Tech. and Ph.D. from Cochin University of Science and Technology, Cochin-22, India. At present, she is working as Professor in the Department of Electronics, Cochin University of Science and Technology. She has to her credit, more than 80 research papers, in various research fields, published in International and National journals and conferences. Her areas of interest include digital signal / image processing, bio medical image processing, super resolution, content based image retrieval, genomic signal processing, etc.

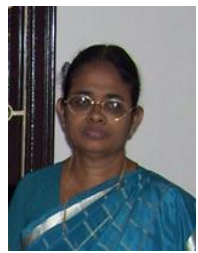